\documentclass[sigconf, natbib]{acmart}
\setcitestyle{numbers,sort&compress,comma,square}
\usepackage{libertine}
\usepackage{booktabs}
\usepackage{hyperref}
\usepackage{graphicx}
\usepackage{amssymb}
\usepackage{amsmath}
\usepackage{subfig}
\usepackage{xspace}
\usepackage{url}
\usepackage{multirow}
\usepackage {tikz}
\usetikzlibrary {positioning}
\definecolor {processblue}{cmyk}{0.96,0,0,0}
\usepackage{caption}
\usepackage{makecell}
\usetikzlibrary{fit}
\usetikzlibrary{shapes.misc}
\usepackage{color}
\usepackage{algorithm}
\usepackage[noend]{algpseudocode}

\makeatletter
\newcommand{\setalglineno}[1]{%
  \setcounter{ALG@line}{\numexpr#1-1}}
\makeatother

\newcommand{\alg}{\textsc{KATE}\xspace}
\newcommand{\vect}[1]{\boldsymbol{#1}}
\newcommand{\bx}{\vect{x}}
\newcommand{\hbx}{\hat{\bx}}
\newcommand{\hx}{\hat{x}}
\newcommand{\bv}{\vect{v}}
\newcommand{\bb}{\vect{b}}
\newcommand{\bc}{\vect{c}}
\newcommand{\bz}{\vect{z}}
\newcommand{\hbz}{\hat{\bz}}
\newcommand{\bW}{\vect{W}}
\newcommand{\setR}{{\mathord{\mathbb R}}}

\tikzset{cross/.style={cross out, draw=black, minimum size=2*(#1-\pgflinewidth), inner sep=0pt, outer sep=0pt},
cross/.default={1pt}}
\tikzset{%
  every neuron/.style={
    circle,
    draw,
    minimum size=.5cm
  },
  neuron missing/.style={
    draw=none, 
    scale=3,
    text height=0.2cm,
    execute at begin node=\color{black}$\vdots$
  },
}

\copyrightyear{2017}
\acmYear{2017}
\setcopyright{acmlicensed}
\acmConference{KDD '17}{August 13-17, 2017}{Halifax, NS, Canada}
\acmPrice{15.00}
\acmDOI{10.1145/3097983.3098017}
\acmISBN{978-1-4503-4887-4/17/08}

\begin{document}

\title{KATE: K-Competitive Autoencoder for Text}


\author{Yu Chen}
\affiliation{%
    \institution{Rensselaer Polytechnic Institute}
    \city{Troy}
    \state{NY, USA}
    \postcode{12180}}
\email{cheny39@rpi.edu}

\author{Mohammed J. Zaki}
\orcid{0000-0003-4711-0234}
\affiliation{%
    \institution{Rensselaer Polytechnic Institute}
    \city{Troy}
    \state{NY, USA}
    \postcode{12180}}
\email{zaki@cs.rpi.edu}

\begin{abstract}

Autoencoders have been successful in learning meaningful representations
from image datasets. However, their performance on text datasets has not
been widely studied. Traditional autoencoders tend to learn possibly
trivial representations of text documents due to their confounding
properties such as high-dimensionality, sparsity and power-law word
distributions. In this paper, we propose a novel k-competitive
autoencoder, called \alg, for text documents. Due to the competition
between the neurons in the hidden layer, each neuron becomes specialized
in recognizing specific data patterns, and overall the model can learn
meaningful representations of textual data. A comprehensive set of
experiments show that \alg can learn better representations than
traditional autoencoders including denoising, contractive, variational,
and k-sparse autoencoders. Our model also outperforms deep generative
models, probabilistic topic models, and even word representation models
(e.g., Word2Vec) in terms of several downstream tasks such as document
classification, regression, and retrieval. 

\end{abstract}

\ccsdesc[500]{Information systems~Data mining}
\ccsdesc[500]{Information systems~Document representation}
\ccsdesc[300]{Information systems~Retrieval tasks and goals}
\ccsdesc[500]{Computing methodologies~Neural networks}
\ccsdesc[300]{Computing methodologies~Information extraction}

\keywords{Representation Learning, Autoencoders, Competitive Learning, Text Analytics} 

\maketitle

\vspace{-0.1in}
\section{Introduction}

An autoencoder is a neural network which can automatically learn data
representations by trying to reconstruct its input at the output layer. 
Many variants of autoencoders have been proposed
recently~\citep{vincent2010stacked, rifai2011contractive, kingma2013auto,
    makhzani2015adversarial, makhzani2013k, makhzani2015winner}. While
autoencoders have been successfully applied to learn meaningful
representations on image datasets (e.g., MNIST~\citep{lecun1998gradient}, CIFAR-10~\citep{krizhevsky2009learning}),
their performance on text datasets has not been widely studied.
Traditional autoencoders are susceptible to learning trivial
representations for text documents. As noted by Zhai and
Zhang~\citep{zhai2015semisupervised}, the reasons include that fact that
textual data is extremely high dimensional and sparse. The vocabulary
size can be hundreds of thousands while the average fraction of zero
entries in the document vectors can be very high (e.g., 98\%). Further, textual
data typically follows power-law word distributions. That is,
low-frequency words account for most of the word occurrences. Traditional
autoencoders always try to reconstruct each dimension of the input
vector on an equal footing, which is not quite appropriate for textual
data.

Document representation is an interesting and challenging task which is
concerned with representing textual documents in a vector space,
and it has various applications in text processing, retrieval and
mining. There are two major approaches to represent documents: 1) {\bf
    Distributional Representation} is based on the hypothesis that
linguistic terms with similar distributions have similar meanings. These
methods usually take advantage of the co-occurrence and context information
of words and documents, and each dimension of the document vector
usually represents a specific semantic meaning (e.g., a topic). Typical
models in this category include Latent Semantic Analysis
(LSA)~\citep{deerwester1990indexing}, probabilistic LSA
(pLSA)~\citep{hofmann1999probabilistic} and Latent Dirichlet Allocation
(LDA)~\citep{blei2003latent}. 2) {\bf Distributed Representations}
encode a document as a compact, dense and lower dimensional vector
with the semantic meaning of the document distributed along the
dimensions of the vector. 
Many neural network-based
distributed representation models \citep{larochelle2012neural,
    srivastava2013modeling, maaloe2015deep, cao2015novel,
    miao2015neural} have been proposed and shown to be able to learn
better representations of documents than distributional representation
models.

In this paper, we try to overcome the weaknesses of traditional
autoencoders when applied to textual data. We propose a novel
autoencoder called \alg (for {\bf K}-competitive {\bf
    A}utoencoder for {\bf TE}xt), which relies on competitive
learning among the autoencoding neurons. In the feedforward phase, only
the most competitive $k$ neurons in the layer fire and those $k$
``winners'' further incorporate the aggregate activation potential of
the remaining inactive neurons. As a result, each hidden neuron becomes
better at recognizing specific data patterns and the overall model can
learn meaningful representations of the input data. After training the
model, each hidden neuron is distinct from the others and no competition
is needed in the testing/encoding phase. We conduct comprehensive experiments
qualitatively and quantitatively to evaluate \alg and to demonstrate the
effectiveness of our model.  We compare \alg with traditional
autoencoders including basic autoencoder, denoising
autoencoder~\citep{vincent2010stacked}, contractive
autoencoder~\citep{rifai2011contractive}, variational
autoencoder~\citep{kingma2013auto}, and k-sparse
autoencoder~\citep{makhzani2013k}. We also compare with deep generative
models~\citep{maaloe2015deep},
neural autoregressive \citep{larochelle2012neural} and
variational inference~\citep{miao2015neural} models,
probabilistic topic models such as LDA~\citep{blei2003latent}, and
word representation models such as
Word2Vec~\citep{mikolov2013distributed} and
Doc2Vec~\citep{le2014distributed}. \alg achieves state-of-the-art
performance across various datasets on several downstream tasks like
document classification, regression and retrieval.

\section{Related Work}

\textbf{Autoencoders.} The basic autoencoder is a shallow neural network
which tries to reconstruct its input at the output layer. An autoencoder
consists of an encoder which maps the input $\bx$ to the hidden layer: 
$\bz = g(\bW\bx + \bb)$ and a decoder which reconstructs the input 
as: $\hbx = o(\bW'\bz + \bc)$; here $\bb$ and $\bc$ are bias terms, 
$\bW$ and $\bW'$ are input-to-hidden and hidden-to-output layer weight
matrices, and $g$ and $o$ are activation functions.
Weight tying (i.e., setting $\bW' = \bW^T$) is
often used as a regularization method to avoid overfitting. While plain
autoencoders, even with perfect reconstructions, usually only extract
trivial representations of the data, more
meaningful representations can be obtained by adding appropriate regularization to the
models. Following this line of reasoning, many variants of autoencoders
have been proposed recently \citep{vincent2010stacked,
    rifai2011contractive, kingma2013auto, makhzani2015adversarial,
    makhzani2013k, makhzani2015winner}. The denoising autoencoder
(DAE) \citep{vincent2010stacked} inputs a
corrupted version of the data while the output is still compared with the
original uncorrupted data, allowing the model 
to
learn patterns useful for denoising. The contractive
autoencoder (CAE) \citep{rifai2011contractive}
introduces the Frobenius norm of the Jacobian matrix 
of the encoder activations into the regularization
term.
When the Frobenius norm is 0, 
the model is extremely invariant to perturbations of input data, 
which is thought as good. 
The variational autoencoder (VAE) \citep{kingma2013auto} is
a generative model inspired by variational inference whose encoder 
$q_\phi(\bz|\bx)$ approximates the intractable true posterior
$p_\theta(\bz|\bx)$,
and the decoder $p_\theta(\bx|\bz)$ is a data generator. 
The k-sparse autoencoder (KSAE) \citep{makhzani2013k} 
explicitly
enforces sparsity by only keeping the $k$ highest activities in the
feedforward phase.

We notice that most of the successful applications of autoencoders
are on image data, while only a few have attempted to apply autoencoders
on textual data. Zhai and Zhang~\citep{zhai2015semisupervised} have argued that
traditional autoencoders, which perform well on image data, are less
appropriate for modeling textual data due to the problems of
high-dimensionality, sparsity and power-law word distributions.
They proposed a semi-supervised autoencoder which
applies a weighted loss function where the weights are learned by a
linear classifier to overcome some of these problems. 
Kumar and D'Haro~\citep{kumar2015deepae} found that all the topics
extracted from the autoencoder were dominated by the most frequent words
due to the sparsity of the input document vectors. Further, they
found that adding sparsity and selectivity penalty terms helped alleviate
this issue to some extent.

\textbf{Deep generative models.} Deep Belief Networks (DBNs) are
probabilistic graphical models which learn to extract a deep
hierarchical representation of the data. The top 2 layers of DBNs form a
Restricted Boltzmann Machine (RBM) and other layers form a sigmoid
belief network. A relatively fast greedy layer-wise pre-training
algorithm \citep{hinton2006fast, hinton2006reducing} is applied to train
the model. Maaloe et al.~\citep{maaloe2015deep} showed that DBNs can be
competitive as a topic model.
DocNADE \citep{larochelle2012neural} is a neural autoregressive topic
model that estimates the probability of observing a new word in a given
document given the previously observed words. It can be used for
extracting meaningful representations of documents. It has been shown to
outperform the Replicated Softmax model~\citep{hinton2009replicated} which
is a variant of RBMs for document modeling. 
Srivastava et al.~\citep{srivastava2013modeling} 
introduced a type of Deep Boltzmann Machine
(DBM) that is suitable for extracting distributed semantic
representations from a corpus of documents; an Over-Replicated Softmax
model was proposed to overcome the apparent difficulty of training a DBM.
NVDM \citep{miao2015neural} is a neural variational inference model for
document modeling inspired by the variational autoencoder.

\textbf{Probabilistic topic models.} Probabilistic topic models, such as
probabilistic Latent Semantic Analysis (pLSA) and Latent Dirichlet
Allocation (LDA) have been extensively
studied \citep{hofmann1999probabilistic, blei2003latent}. Especially for
LDA, many variants have been proposed for non-parametric
learning \citep{teh2012hierarchical, blei2010nested},
sparsity \citep{wang2009decoupling, eisenstein2011sparse, zhu2012sparse}
and efficient inference \citep{teh2007collapsed, canini2009online}. Those
models typically build a generative probabilistic model using the bag-of-words representation of the documents.

\textbf{Word representation models.} Distributed representations of
words in a vector space can capture semantic meanings of words and help
achieve better results in various downstream text analysis tasks.
Word2Vec~\citep{mikolov2013distributed} and
Glove~\citep{pennington2014glove} are state-of-the-art
word representation models. Pre-training word embeddings on a large
corpus of documents and applying learned word embeddings in downstream
tasks has been shown to work well in practice
\citep{kim2014convolutional, das2015gaussian, nguyen2015improving}.
Doc2Vec~\citep{le2014distributed} was inspired by Word2Vec and can directly
learn vector representations of paragraphs and documents.
NTM~\citep{cao2015novel},
which also uses pre-trained word embeddings,
is a neural topic model where the representations
of words and documents are combined into a uniform framework.

With this brief overview of existing work, we now turn to our
competitive autoencoder approach for text documents.

\section{K-Competitive Autoencoder}

Although the objective of an autoencoder is to minimize the
reconstruction error, our goal is to extract meaningful features from
data. Compared with image data, textual data is more challenging for
autoencoders since it is typically high-dimensional, sparse and
has power-law word distributions. When examining the features extracted by
an autoencoder, we observed that they were not distinct from one
another. That is, many neurons in the hidden layer shared similar groups
of input neurons (which typically correspond to the most frequent words) 
with whom they had the strongest connections. 
We hypothesized that the autoencoder greedily learned relatively trivial features
in order to reconstruct the input. 

To overcome this
drawback, our approach guides the autoencoder to focus on important 
patterns in the data by adding constraints in the training
phase via mutual competition.
In competitive learning, neurons compete for the right
to respond to a subset of the input data and as a result, the
specialization of each neuron in the network is increased. Note that the
specialization of neurons is exactly what we want for an autoencoder,
especially when applied on textual data. By introducing competition into
an autoencoder, we expect each neuron in the hidden layer to take
responsibility for recognizing different patterns within the input data.
Following this line of reasoning, we propose the k-competitive
autoencoder, \alg, as described below.

\subsection{Training and Testing/Encoding}

\begin{algorithm}[!h]
\caption{\alg: K-competitive Autoencoder}
\label{alg:kcae}
\begin{algorithmic}[1]

\Procedure{Training}{}
\State Feedforward step: $\bz = tanh(\bW\bx + \bb)$
\State Apply k-competition: $\hbz = \text{k-competitive\_layer}(\bz)$
\State Compute output: $\hbx = sigmoid(\bW^T\hbz + \bc)$
\State Backpropagate error (cross-entropy) and iterate
\EndProcedure

\item[]
\setalglineno{1}
\Procedure{Encoding}{}
\State Encode input data: $\bz = tanh(\bW\bx + \bb)$
\EndProcedure
\end{algorithmic}
\end{algorithm}

The pseudo-code for our k-competitive autoencoder \alg is shown in
Algorithm~\ref{alg:kcae}. \alg is a shallow autoencoder with 
a (single) competitive hidden layer, with each neuron
competing for the right to respond to a given set of input patterns.
Let $\bx \in \setR^d$ be a $d$-dimensional input vector, which is also
the desired output vector, and 
let $h_1, h_2, ..., h_m$ be the $m$ hidden neurons.
Let $\bW \in \setR^{d \times m}$ be the weight matrix linking the input layer to
the hidden layer neurons, and let $\bb\in\setR^m$ and $\bc \in \setR^d$ 
be the bias terms for
the hidden and output neurons, respectively. 
Let $g$ be an activation function; two typical
functions are $tanh(x) = \frac{e^{2x}-1}{e^{2x}+1}$ and 
$sigmoid(x) = \frac{1}{1+e^{-x}}$.
In each feed-forward step,
the activation potential at the hidden neurons is then given as
$\bz = g(\bW\bx+\bb)$, whereas the activation potential at the output
neurons is given as $\hbx = g(\bW^T\bz+\bc)$. Thus,
the hidden-to-output weight matrix is simply $\bW^T$, being an instance
of {\em weight tying}.

In \alg, 
we represent each input text document as a log-normalized word
count vector $\bx \in \setR^d$ where each dimension is represented as
$$x_i = \frac{log (1 + n_i)}{\max_{i \in V}{log (1+ n_i)}}, \text{ for } i \in V$$
where $V$ is the vocabulary and $n_i$ is the count of word $i$
in that document. Let $\hbx$ be the output of \alg on a given input
$\bx$. 
We use the binary cross-entropy as the loss function, which is
defined as 
$$l(\bx, \hbx) = -\sum_{i \in V}{x_i log(\hx_i) + (1-x_i) log(1-\hx_i)}$$
where $\hx_i$ is the reconstructed value for $x_i$.

Let $H$ be some subset of hidden neurons; define
the {\bf energy} of $H$ as the total activation potential for $H$, 
given as: $E(H) = \sum_{h_i \in H} |z_i|$, i.e., sum of the absolute
values of the activations for neurons in $H$.
In \alg, in the feedforward phase, 
after computing the activations $\bz$ for a given input $\bx$,  
we select the
most competitive $k$ neurons  
as the ``winners'' while the remaining ``losers'' are suppressed (i.e.,
made inactive).
However, in order to 
compensate for the loss of energy from the loser neurons, and
to make the competition among neurons more pronounced, we amplify and reallocate
that energy among the winner neurons.

\begin{algorithm}[!h]
\caption{K-competitive Layer}\label{alg:kcl}
\begin{algorithmic}[1]

\Function{k-competitive-layer}{$\bz$}

\State sort positive neurons in ascending order $z^+_1 ... z^+_P$
\State sort negative neurons in descending order $z^-_1 ... z^-_N$


\If {$P - \lceil k/2 \rceil > 0$}

\State $E_{pos} = \sum_{i=1}^{P - \lceil k/2 \rceil}{z^+_i}$
 \For{\texttt{$i = P - \lceil k/2 \rceil + 1, ..., P$}}
        \State \texttt{$z^+_i := z^+_i + \alpha \cdot E_{pos}$}
 \EndFor
 
 \For{\texttt{$i = 1,...,P - \lceil k/2 \rceil$}}
        \State \texttt{$z^+_i := 0$}
 \EndFor

\EndIf

    
\If {$N - \lfloor k/2 \rfloor > 0$}

\State $E_{neg} = \sum_{i=1}^{N - \lfloor k/2 \rfloor}{z^-_i}$
 \For{\texttt{$i = N - \lfloor k/2 \rfloor + 1, ..., N$}}
        \State \texttt{$z^-_i := z^-_i + \alpha \cdot E_{neg}$}
 \EndFor
 
 \For{\texttt{$i = 1,...,N - \lfloor k/2 \rfloor$}}
        \State \texttt{$z^-_i := 0$}
 \EndFor

\EndIf

\State \Return updated $z^+_1...z^+_P, z^-_1 ... z^-_N$

\EndFunction

\end{algorithmic}
\end{algorithm}

\begin{figure}[!h]
\vspace{-0.15in}
\centering

\begin{tikzpicture}[x=1.3cm, y=.75cm, >=stealth]

 \foreach \m [count=\y] in {1,missing,2}
  \node [every neuron/.try, neuron \m/.try ] (input-\m) at (-1,-\y*1.25) {};
  
\foreach \m [count=\y] in {1,2, 3, 4, 5,6}
  \node [every neuron/.try, neuron \m/.try ] (hidden-\m) at (2,2-\y*1.25) {};

\foreach \m [count=\y] in {1,2}{
  \node [every neuron/.try, neuron \m/.try ] (adder-\m) at (1,-1-\y*1.25) {};
  \draw (1,-1-\y*1.25) node[cross=5pt,rotate=45]{};
}
 
 \foreach \m [count=\y] in {1,missing,2}
  \node [every neuron/.try, neuron \m/.try ] (output-\m) at (4,-\y*1.25) {};

   \node [above] at (input-1.north) {$i_1$};
 \node [above] at (input-2.north) {$i_d$};
   
   \node [above] at (output-1.north) {$o_1$};
 \node [above] at (output-2.north) {$o_d$};
  
 \node [above] at (hidden-1.north) {$h_1$};
 \node [above] at (hidden-2.north) {$h_2$};
  \node [above] at (hidden-3.north) {$h_3$};
    \node [above] at (hidden-4.north) {$h_4$};
      \node [above] at (hidden-5.north) {$h_5$};
 \node [above] at (hidden-6.north) {$h_6$};
 
  \node [right] at (hidden-1.east) {0.8 \color{red} $+0.3\alpha$};
   \node [right] at (hidden-2.east) {0.2 \color{red} $\rightarrow 0$};
 \node [right] at (hidden-3.east) {0.1 \color{red} $\rightarrow 0$};
 \node [right] at (hidden-4.east) {-0.1 \color{red} $\rightarrow 0$};
 \node [right] at (hidden-5.east) {-0.3 \color{red} $\rightarrow 0$}; 
  \node [right] at (hidden-6.east) {-0.6 \color{red} $-0.4\alpha$};
  
 \node [above] at (adder-1.west) {$PosAdder$\space\space\space\space\space\space\space\space\space\space\space\space\space\space};
 \node [above] at (adder-2.west) {$NegAdder$\space\space\space\space\space\space\space\space\space\space\space\space\space\space};
 \node [above] at (adder-1.north) {\color{red}0.3\space\space\space\space\space};
 \node [above] at (adder-2.north) {\color{red}-0.4\space\space\space\space\space}; 
 
 
  \draw  [->] (input-1) -- (hidden-1);
    \draw  [->] (input-1) -- (hidden-6);
  \draw  [->] (input-2) -- (hidden-1);
    \draw  [->] (input-2) -- (hidden-6);
    	     
 \draw  [->] (adder-1) -- (hidden-1) node [midway, above] (TextNode) {$\alpha$};
  \draw  [->] (adder-2) -- (hidden-6) node [midway, below] (TextNode) {$\beta$};
   \draw  [->] (hidden-2) -- (adder-1);
    \draw  [->] (hidden-3) -- (adder-1);
     \draw  [->] (hidden-4) -- (adder-2);
      \draw  [->] (hidden-5) -- (adder-2);

 \draw  [->] (hidden-1) -- (output-1);
 \draw  [->] (hidden-1) -- (output-2);
  \draw  [->] (hidden-6) -- (output-1);
   \draw  [->] (hidden-6) -- (output-2);
   
\node[draw=red, dotted, fit=(hidden-1) (hidden-2) (hidden-3) (hidden-4) (hidden-5) (hidden-6)] {};
\node [align=center, above] at (1*-1,-7.2) {Input \\ layer};
\node [align=center, above] at (1*2,-7.2) {k-competitive \\ layer};
\node [align=center, above] at (1*4,-7.2) {Output \\ layer};
\end{tikzpicture}
\vspace{-0.1in}
\caption{Competitions among neurons.  
Input and hidden neurons, and hidden and output neurons
are fully connnected, but we omit these to avoid clutter.}
\label{fig:competition}
\vspace{-0.2in}
\end{figure}
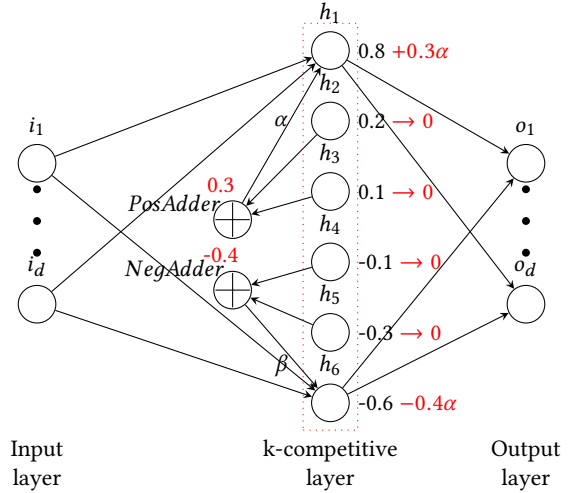

\alg uses {\em tanh} activation function for the  
k-competitive hidden layer. We divide these 
neurons into positive and negative neurons based on their activations. 
The most competitive 
$k$ neurons are those that have the largest
absolute activation values. 
However, we select the $\lceil k/2 \rceil$ largest positive activations
as the positive winners, and reallocate the energy of the remaining
positive loser neurons among the winners using an $\alpha$ amplification
connection, where $\alpha$ is a hyperparameter.
Finally, we set the
activations of all losers to zero. 
Similarly,
the $\lfloor k/2 \rfloor$ lowest negative
activations are the negative winners, and they incorporate the amplified energy
from the negative loser neurons, as detailed in 
Algorithm~\ref{alg:kcl}. 
We argue that the $\alpha$ amplification connections are a critical
component in the k-competitive layer.
When $\alpha=0$, no gradients will flow through loser neurons, resulting
in a regular k-sparse autoencoder (regardless of the activation functions and k-selection scheme).
When $\alpha > 2/k$, we actually boost the gradient signal flowing
through the loser neurons.
We empirically show that amplification helps improve the autoencoder model (see
Sec.\ \ref{sec:mscd} and \ref{sec:effects_of_params}).
As an example, consider 
Figure~\ref{fig:competition}, which shows an
example feedforward step for $k=2$. Here, $h_1$ and $h_6$
are the positive and negative winners, respectively, since the
absolute activation potential for $h_1$ is $|z_1| = 0.8$, and for $h_6$
it is $|z_6| = 0.6$.
The positive winner $h_1$ takes away
the energy from the positive losers $h_2$ and $h_3$, which is
$E(\{h_2,h_3\}) = 0.2+0.1 = 0.3$.
Likewise, the negative winner $h_6$ takes away
the energy from the negative losers $h_4$ and $h_5$, which is 
$-E(\{h_4, h_5\}) = -(|-0.1| + |-0.3|) = -0.4$.
The hyperparameter $\alpha$ governs how the energy from the loser
neurons is incorporated into
the winner neurons, for both positive and negative cases.
That is $h_1$'s net activation becomes $z_1 = 0.8+0.3\alpha$, and $h_6$'s
net activation is $z_6 = -0.6-0.4\alpha$. The rest of the neurons are
set to zero activation.

Finally, as noted in Algorithm~\ref{alg:kcae} we use weight tying for
the hidden to output layer weights, i.e., we use $\bW^T$ as the weight
matrix, with different biases $\bc$. Also, since the inputs $\bx$ are
non-negative for document representations (e.g., word counts), 
we use the {\em sigmoid}
activation function at the output layer to maintain the non-negativity.
Note that in the back-propagation procedure, 
the gradients will first flow through the winner neurons in the hidden layer and then
the loser neurons via the $\alpha$ amplification connections.
No gradients will flow directly from the output neurons to
the loser neurons since they are made inactive in the feedforward step.

\subsubsection*{Testing/Encoding}

Once the k-competitive network has been trained, we simply encode each
test input as shown in Algorithm~\ref{alg:kcae}.
That is, given a test input $\bx$, we map it to the feature space to obtain
$\bz = tanh(\bW\bx + \bb)$.
No competition is required for the encoding step
since the hidden neurons are well trained to be distinctive from others.
We argue that this is one of the superior features of \alg.

\subsection{Relationship to Other Models}

\subsubsection*{KATE vs.\ K-Sparse Autoencoder}

The k-sparse autoencoder~\citep{makhzani2013k} 
is closely related to our
model, but there are several differences. The k-sparse autoencoder explicitly
enforces sparsity by only keeping the $k$ highest activities at training
time. Then, at testing time, in order to enforce sparsity, only the $\alpha
k$ highest activities are kept where $\alpha$ is a hyperparameter. 
Since its hidden layer uses a linear activation function,
the only non-linearity in the encoder comes 
from the selection of the $k$ highest activities.
Instead of focusing on sparsity, our
model focuses on competition 
to drive each hidden neuron to be distinct from the others.
Thus, at testing time, no competition is needed.
The non-linearity in KATE's encoding
comes from the {\em tanh} activation function and the winner-take-all
operation (i.e., top k selection and amplifying energy reallocation).

It is important to note that for the k-sparse autoencoder,
too much sparsity (i.e., low $k$) can cause the
so-called ``dead'' hidden neurons problem, 
which can prevent gradient back-propagation from 
adjusting the weights of these ``dead'' hidden neurons.
As mentioned in the original paper, the model is prone to
behaving in a manner similar to k-means clustering.
That is, in the first few epochs, it will greedily assign 
individual hidden neurons to groups of training cases
and these hidden neurons will be re-enforced but other
hidden neurons will not be adjusted in subsequent epochs.
In order to address this problem, scheduling the sparsity level over epochs
was suggested. However, by design 
our approach does not suffer from this problem
since the gradients will still flow through the loser neurons via
the $\alpha$ amplification connections in the k-competitive layer.

\subsubsection*{KATE vs.\ K-Max Pooling}
Our proposed k-competitive operation
is also reminiscent of the k-max pooling
operation~\citep{blunsom2014convolutional}
applied in convolutional neural networks.
We can intuitively regard k-max pooling
as a global feature sampler which selects 
a subset of k maximum neurons in the previous convolutional layer
and uses only the selected subset of neurons in the following layer.
Unlike our k-competitive approach, the objective of k-max pooling
is to reduce dimensionality and introduce feature invariance 
via this downsampling operation.

\subsubsection*{KATE as a Regularized Autoencoder}

We can also regard our model as a special case of a fully competitive
autoencoder where all the neurons in the hidden layer are fully
connected with each other and the weights on the connections between
them are fully trainable. The difference is that we restrict the
architecture of this competitive layer by using a positive adder and a
negative adder to constrain the energy, which serves as a
regularization method. 


\section{Experiments}

In this section, we evaluate our k-competitive autoencoder
model on various datasets and downstream text analytics tasks to gauge
its effectiveness in 
learning meaningful representations in different situations.
All experiments were performed on a machine with a
1.7GHz AMD Opteron 6272 Processor, with 264G RAM.
Our model, \alg, was implemented in Keras
(\url{github.com/fchollet/keras}) which is a high-level neural networks library, written in Python.
The source code for KATE is available at
\textbf{\url{github.com/hugochan/KATE}}.

\begin{table}[!h]
    \small
  \centering
  \renewcommand{\arraystretch}{1.0}
	\begin{tabular}{| p{1.6cm}| p{1.3cm}| p{1.3cm}| p{1.3cm}| p{1.4cm}|}
 
    \hline
     dataset  & 20 news & reuters & wiki10+ & mrd \\
    \hline
   train.size & 11,314 & 554,414 & 13,972 & 3,337\\ 
   test.size & 7,532 & 250,000 & 6,000 & 1,669 \\
   valid.size & 1,000 & 10,000 & 1,000 & 300\\
   vocab.size & 2,000  & 5,000 &  2,000 & 2,000  \\ 
   avg.length & 93 & 112 & 1,299  & 124 \\
   classes/vals & 20 & 103 & 25 & $[0,1]$\\
   task & class \& DR & MLC & MLC & regression \\ 
  \hline\end{tabular}
  \caption{Datasets: Tasks include classification (class), regression, multi-label
      classification (MLC), and document retrieval (DR).}
  \label{table:datasets}
  \vspace{-0.2in}
\end{table}

\subsection{Datasets}
For evaluation, we use datasets that have been widely used in
previous studies~\citep{Lang95, lewis2004rcv1, zubiaga2009getting, pang2002thumbs, pang2004sentimental, pang2005seeing}.
Table~\ref{table:datasets} provides statistics of the different
datasets used in our experiments. It lists the training, testing and
validation (a subset of training) set sizes, the size of the vocabulary,
average document length, the number of classes (or values for regression), 
and the various
downstream tasks we perform on the datasets.

The 20 Newsgroups~\citep{Lang95} (\url{www.qwone.com/~jason/20Newsgroups})
data consists of 18846 documents, which are partitioned (nearly) evenly
across 20 different newsgroups. Each document belongs to exactly one newsgroup.
The corpus is divided by date into training (60\%) and testing (40\%) sets. 
We follow the preprocessing steps utilized in previous
work~\citep{larochelle2012neural, srivastava2013modeling, miao2015neural}. 
That is, after removing stopwords and stemming, 
we keep the most frequent 2,000 words in the training set as the vocabulary. 
We use this dataset to show that our model
can learn meaningful representations for classification
and document retrieval tasks.

The Reuters
RCV1-v2 dataset~\citep{lewis2004rcv1} (\url{www.jmlr.org/papers/volume5/lewis04a})
contains 804,414 newswire articles, where each
document typically has multiple (hierarchical) topic labels. The total
number of topic labels is 103. The dataset already comes preprocessed
with stopword removal and stemming. 
We randomly split the corpus into 554,414 training and 25,000 test cases and
keep the most frequent 5,000 words in the training dataset as the vocabulary. 
We perform multi-label classification on this dataset.

The Wiki10+ dataset~\citep{zubiaga2009getting}
(\url{www.zubiaga.org/datasets/wiki10+/}) comprises English Wikipedia articles with at least 10
annotations on delicious.com. Following the steps of Cao et
al.~\citep{cao2015novel},
we only keep the 25 most frequent social tags and those documents
containing any of these tags. 
After removing stopwords and stemming, 
we randomly split the corpus into 13,972 training and 6,000
test cases and keep the most frequent 2,000 words in the training set as
the vocabulary for use in 
multi-label classification.

The Movie review data (MRD)~\citep{pang2002thumbs, pang2004sentimental, pang2005seeing}
(\url{www.cs.cornell.edu/people/pabo/movie-review-data/})
contains a collection of movie-review documents, with a 
numerical rating score in the interval $[0,1]$.
After removing stopwords and
stemming, we randomly split the corpus into 3,337 training and 1,669
test cases and keep the most frequent 2,000 words in the training set as
the vocabulary. We use this dataset for regression, i.e., predicting the
movie ratings.

Note that among the above datasets, only the 20 Newsgroups dataset is
balanced, whereas
both the Reuters and Wiki10+ datasets are highly imbalanced in terms of
class labels.

\subsection{Comparison with Baseline Methods}

We compare our k-competitive autoencoder \alg with a wide range of other
models including various types of autoencoders, topic models, belief
networks and word representation models, as listed below.

\textbf{LDA}~\citep{blei2003latent}: a directed graphical model which models a document as a
mixture of topics and a topic as a mixture of words. Once trained, each
document can be represented as a topic proportion vector on the topic
simplex. We used the gensim~\citep{rehurek_lrec} LDA implementation in our experiments.

\textbf{DocNADE}~\citep{larochelle2012neural}: a neural autoregressive topic model that can be used
for extracting meaningful representations of documents.
The implementation is available at
\url{www.dmi.usherb.ca/~larocheh/code/DocNADE.zip}.



\textbf{DBN}~\citep{maaloe2015deep}: a direct acyclic graph whose top two layers
form a restricted Boltzmann machine. We use the implementation
available at
\url{github.com/larsmaaloee/deep-belief-nets-for-topic-modeling}.

\textbf{NVDM}~\citep{miao2015neural}: a neural variational inference model for document
modeling. The authors have not released the source code,
but we used an open-source implementation at
\url{github.com/carpedm20/variational-text-tensor-flow}.

\textbf{Word2Vec}~\citep{mikolov2013distributed}: a model in which 
each document is represented as the average of the word
embedding vectors for that document.
We use Word2Vec$_{pre}$ to denote the version where
we use Google News pre-trained word embeddings which
contain 300-dimensional vectors for 3 million words and phrases. Those
embeddings were trained by state-of-the-art word2vec skipgram model.
On the other hand, we use Word2Vec to denote the version where we train word embeddings
separately on each of our datasets, using the gensim~\citep{rehurek_lrec} implementation.

\textbf{Doc2Vec}~\citep{le2014distributed}: a distributed representation model inspired by Word2Vec
which can directly learn vector representations of documents. 
There are two versions named Doc2Vec-DBOW and Doc2Vec-DM.
We use Doc2Vec-DM in our experiments as it was reported to 
consistently outperform Doc2Vec-DBOW in the original paper.
We used the gensim~\citep{rehurek_lrec} implementation in our experiments.

\textbf{AE}: a plain shallow (i.e., one hidden layer) autoencoder, 
without any competition, 
which can automatically learn data
representations by trying to reconstruct its input at the output layer. 

\textbf{DAE}~\citep{vincent2010stacked}: a denoising autoencoder that
accepts a corrupted
version of the input data while the output is still the original
uncorrupted data. 
In our experiments, we found that masking noise consistently outperforms other two types of
noise, namely Gaussian noise and salt-and-pepper noise.
Thus, we only report the results of using masking noise.
Basically, masking noise perturbs the input by setting a fraction $v$ of the elements $i$ in each input vector as 0.
To be fair and consistent, we use a shallow denoising autoencoder in our experiments.

\textbf{CAE}~\citep{rifai2011contractive}: a contractive autoencoder which 
introduces the Frobenius norm of the Jacobian matrix 
of the encoder activations into the regularization term.

\textbf{VAE}~\citep{kingma2013auto}: a generative autoencoder inspired by
variational inference. 

\textbf{KSAE}~\citep{makhzani2013k}: a competitive autoencoder which explicitly enforces
sparsity by only keeping the $k$ highest activities in the feedforward
phase.

We implemented the AE, DAE, CAE, VAE and KSAE autoencoders on our own,
since their implementations are not publicly available. 

\smallskip
\noindent{\bf Training Details:} For all the autoencoder models
(including AE, DAE, CAE, VAE, KSAE, and \alg), 
we represent each input document as a log-normalized word
count vector, using binary cross-entropy as the loss function
and Adadelta~\citep{zeiler2012adadelta} as the optimizer. Weight
tying is also applied. For CAE and VAE, additional regularization
terms are added to the loss function as mentioned in the original
papers. As for VAE, we use {\em tanh} as the nonlinear activation function
while as for AE, DAE and CAE, {\em sigmoid} is applied.
As for KSAE, we found that omitting sparsity in the testing phase
gave us better results in all experiments.

When training models, we randomly extract a subset of documents from the
training set as a validation set, as noted in Table~\ref{table:datasets}, 
which is used for tuning hyperparameters and early stopping. 
Early stopping is a type of regularization used to 
avoid overfitting when training an iterative algorithm.
We stop training after 5 successive epochs with no improvement on the validation set.
All baseline
models were optimized as recommended in original sources.
For \alg, 
we set $\alpha$ as 6.26, learning rate as 2, batch size as 100 
(for the Reuters dataset) or 50 (for other datasets) and
$k$ as 6 (for the 20 topics case), 32 (for the 128 topics case) or 
102 (for the 512 topics case), as determined from the
validation set.


\subsection{Qualitative Analysis}

In this set of qualitative experiments, we compare the topics generated
by \alg to other representative models 
including AE, KSAE, and LDA.
Even though \alg is not explicitly designed for the purpose of word embeddings,
we compared word representations learned by \alg with the Word2Vec model
to demonstrate that our model can learn semantically meaningful
representations from text.
We evaluate the above models on the 20 Newsgroups data.
Matching the number of classes, the number of topics is set to 
exactly 20 for all models. For both KSAE and \alg, $k$ (the sparsity
level/number of winning neurons) is set as 6.

\begin{table}[!h]
  \vspace{-0.1in}
    \small
 \centering
 \renewcommand{\arraystretch}{1.}	
  \begin{tabular}{|p{1.75cm}|c|c|} \hline
   
    \multirow{4}{.1cm}{\makecell{soc.religion\\.christian}}
    & AE & \makecell{line subject organ articl peopl\\ post time make write good} \\ \cline{2-3}
    & KSAE &  \makecell{peopl origin bottom applic mind\\ subject pad europ role christian} \\  \cline{2-3}
    	& \alg & \makecell{god christian jesu moral rutger\\ bibl exist religion apr christ}\\  \cline{2-3}
    	& LDA & \makecell{god christian jesu church bibl\\ peopl christ man time life}\\ \hline
    
     \multirow{2}{.1cm}{\makecell{sci.crypt}}
     & \alg& \makecell{govern articl law key encrypt\\ clipper chip secur case distribut} \\ \cline{2-3}
    	 & LDA & \makecell{key encrypt chip clipper secur\\ govern system law escrow privaci} \\ \hline
    	 
      \multirow{2}{.1cm}{\makecell{comp.os.ms-\\windows.misc}}
     & \alg & \makecell{ca system univers window\\ problem card file driver drive scsi} \\ \cline{2-3}
    	 & LDA & \makecell{drive card gun control disk\\ scsi system driver hard bu} \\ \hline
    	 
  \end{tabular}
  \caption{Topics learned by various models.}
  \label{table:topics}
  \vspace{-0.2in}
\end{table}

\subsubsection{Topics Generated by Different Models}
Table~\ref{table:topics} shows some topics learned by various models.
As for autoencoders, each topic is represented by the 
10 words (i.e., input neurons) with the strongest connection to that topic (i.e., hidden neuron).
As for LDA, each topic is represented by the 10 most probable words in that topic.
The basic AE is not very good at learning distinctive topics
from textual data. In our experiment, all the topics learned by AE are
dominated by frequent common words like \textit{line},
\textit{subject} and \textit{organ}, which were always the top 3 words in all
the 20 topics. KSAE learns some meaningful words but only alleviates
this problem to some extent, for example, \textit{line},
\textit{subject}, \textit{organ} and \textit{white} still appears as top
4 words in 6 topics. For this reason, the output of AE and KSAE is shown
for only one of the newsgroups ({\em soc.religion.christian}).
On the other hand, we find that \alg generates 20 topics that are
distinct from each other, and which capture the underlying semantics
very well. For example, it associates words such as
\textit{god, christian, jesu, moral, bibl, exist, religion, christ}
under the topic
\textit{soc.religion.christian}. 
It is worth emphasizing that \alg belongs to the class of 
distributed representation models, where
each topic is ``distributed'' among a group of hidden neurons 
(the topics are therefore better interpreted as ``virtual'' topics).
However, we find that \alg can
generate competitive topics compared with LDA, which explicitly infers
topics as mixture of words.

\begin{table*}[!ht]
  \centering
  \renewcommand{\arraystretch}{1.0}
  \begin{tabular}{|p{1.4cm}|c|c|c|c|c|c|c|c|}
    \hline
 Model & \textbf{weapon}  & \textbf{christian} & \textbf{compani} & \textbf{israel} & \textbf{law} & \textbf{hockey} & \textbf{comput} & \textbf{space} \\ \hline

  \multirow{5}{.1cm}{AE}
   &effort & close & hold & cost & made & plane & inform & studi\\ 
   &muslim & test & simpl& isra & live & tie & run & data\\
   &sort  &larg & serv & arab & give & sex & program & answer\\ 
   &america &  result & commit & fear & power & english & base & origin\\ 
   &escap &answer & societi & occupi & reason & intel & author & unit\\ \hline

  \multirow{5}{.1cm}{KSAE}
   &qualiti & god & commit & occupi & back & int & run & data\\ 
   &challeng & power & student& enhanc & govern & cco & inform & process\\
   &tire & lie  &age & azeri & reason & monash & part & answer\\ 
   &7u &  logic & hold & rpi & call & rsa & case & heard\\ 
   &learn &simpl & consist & sleep & answer & pasadena & start & version\\ \hline   

  \multirow{5}{.1cm}{\alg}
   &arm & belief & market & arab & citizen & playoff & scienc & launch\\ 
   &crime & god & dealer& isra & constitut & nhl & dept & orbit\\
   &gun & believ  &manufactur & palestinian & court & team & cs & mission\\ 
   &firearm &  faith & expens & occupi & feder & wing & math & shuttl\\ 
   &handgun &bibl & cost & jew & govern & coach & univ & flight\\ \hline
   
  \multirow{5}{.1cm}{Word2Vec}
   &assault & understand & insur & lebanon & court & sport & engin & launch \\ 
   &militia & belief & feder& isra & prohibit & nhl & colleg & jpl \\
   &possess & believ  &manufactur & lebanes & ban & playoff & umich & nasa \\ 
   &automat &  god & industri & arab & sentenc & winner & subject & moon \\ 
   &gun &truth & pay & palestinian & legitim & cup & perform & gov \\ \hline
   
  \end{tabular}
  \captionsetup{justification=centering}
  \caption{Five nearest neighbors in the word representation space on 20
      Newsgroups dataset.}
  \label{table:similar_words}
  \vspace{-0.2in}
\end{table*}

\begin{figure*}[!ht]
  \vspace{-0.1in}
  \centerline{
  \subfloat[t][AE]{%
    \includegraphics[width=1.75in]{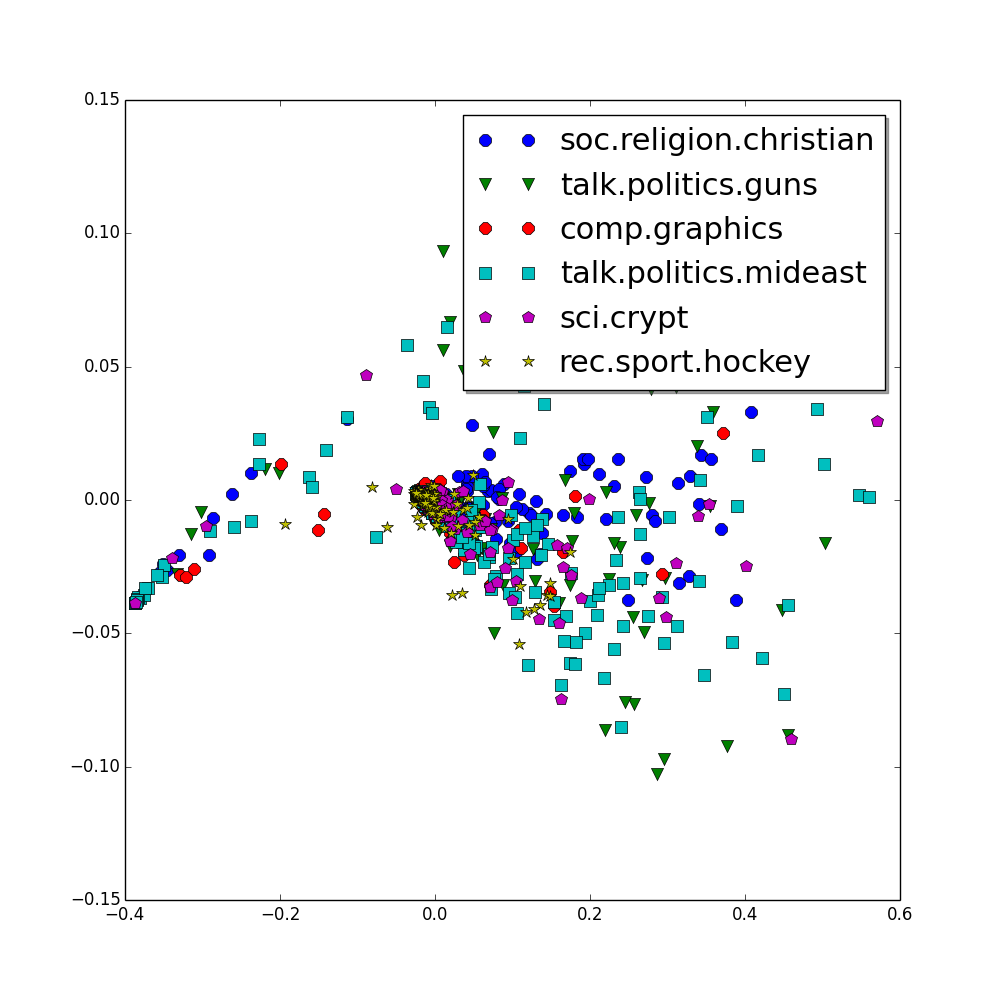}
	\label{fig:PCA_AE}
	}
  \subfloat[t][KSAE]{%
    \includegraphics[width=1.75in]{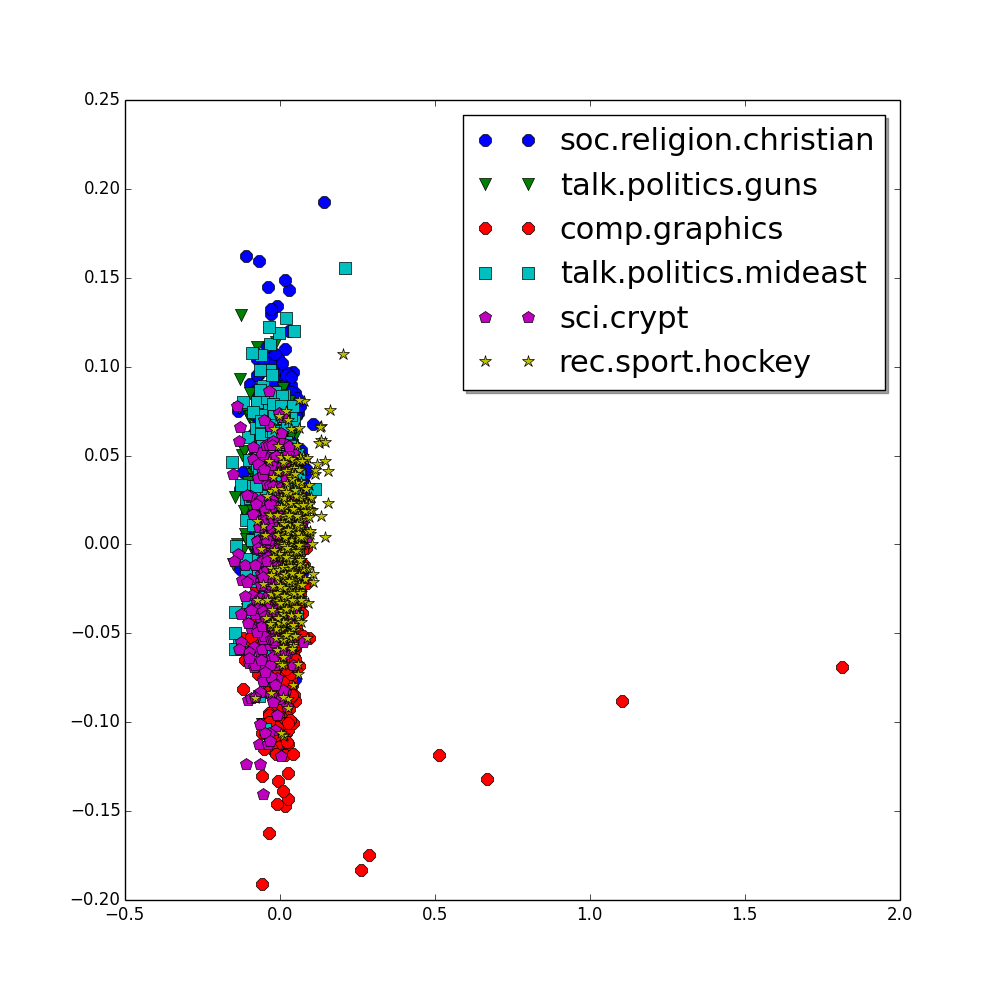}
    \label{fig:PCA_KSAE}
    }
  \subfloat[t][LDA]{%
    \includegraphics[width=1.75in]{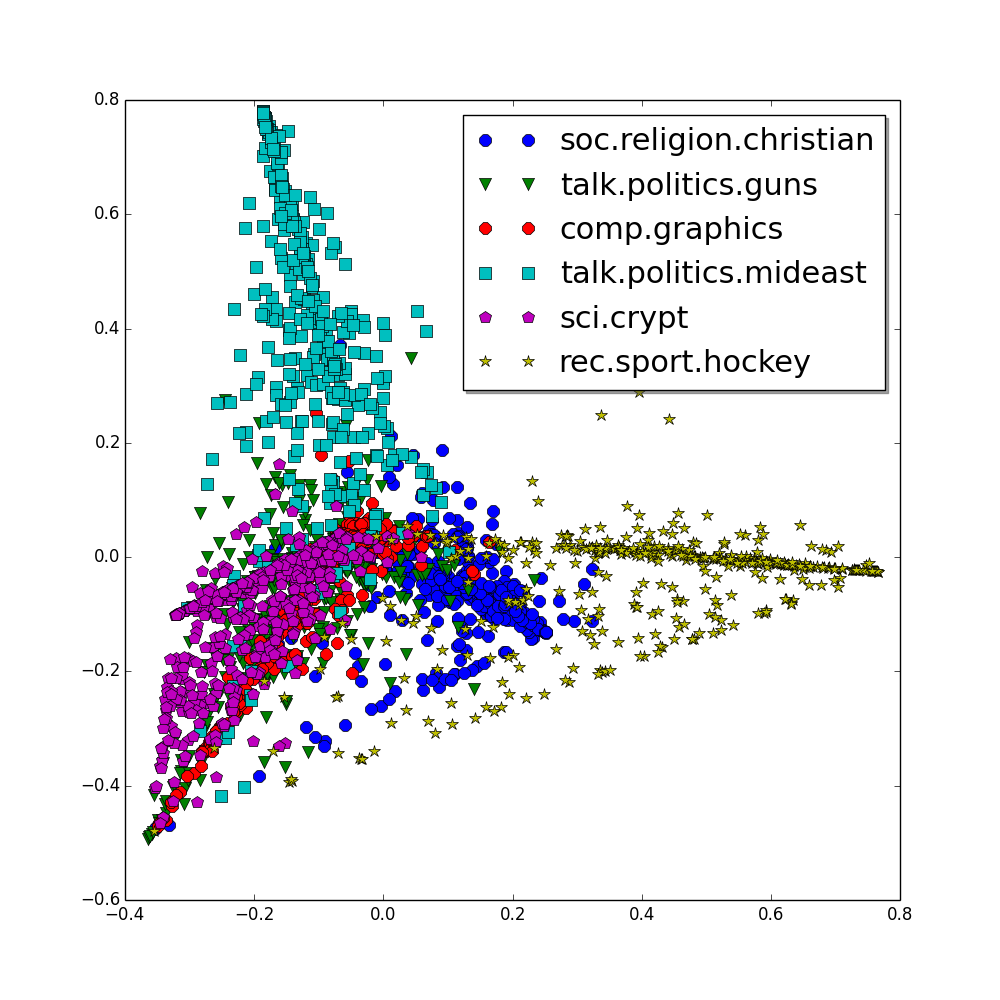}
    \label{fig:PCA_LDA}
    }
  \subfloat[t][\alg]{%
    \includegraphics[width=1.75in]{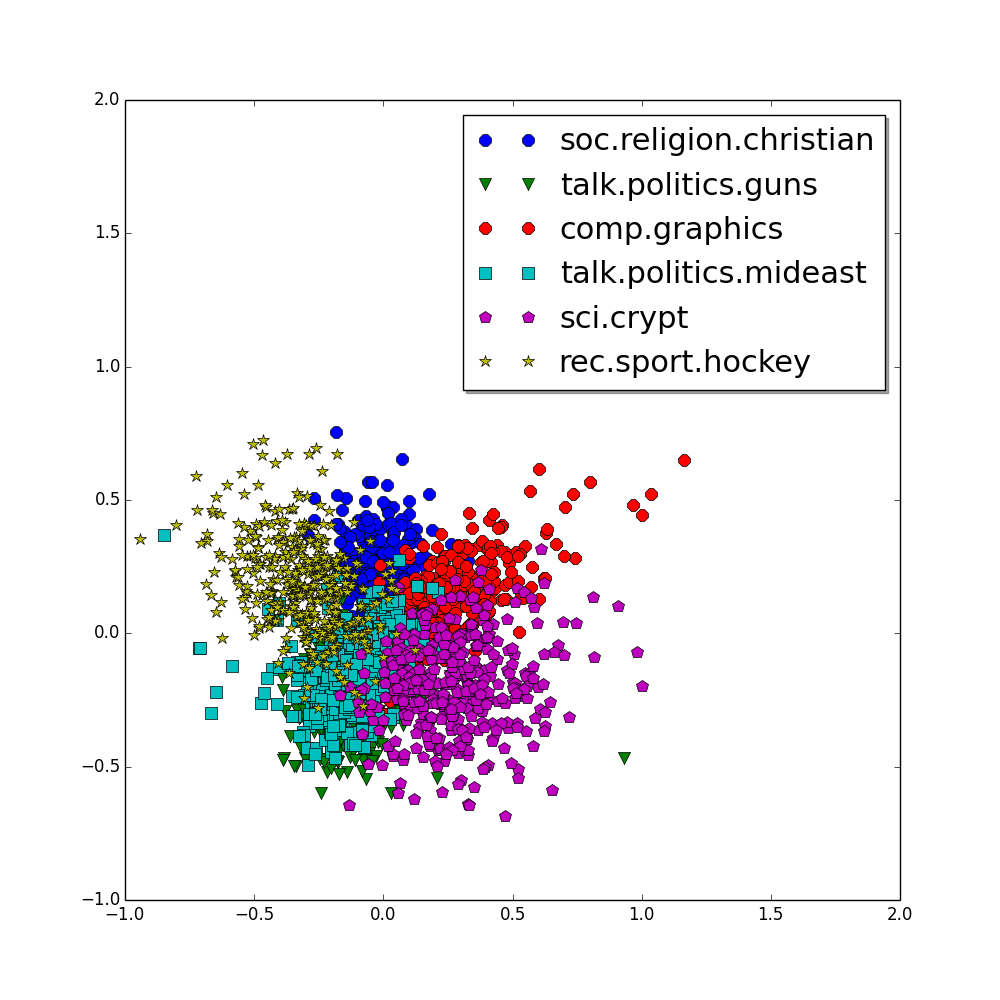}
    \label{fig:PCA_KATE}
   }}
  \vspace{-0.1in}
  \caption{PCA on the 20-dimensional document vectors from 20
      Newsgroups dataset.}
  \label{fig:PCA}
  \vspace{-0.2in}
\end{figure*}

\begin{figure*}[!ht]
  \vspace{-0.1in}
  \centerline{
  \subfloat[t][AE]{%
    \includegraphics[width=1.75in, height=1.5in]{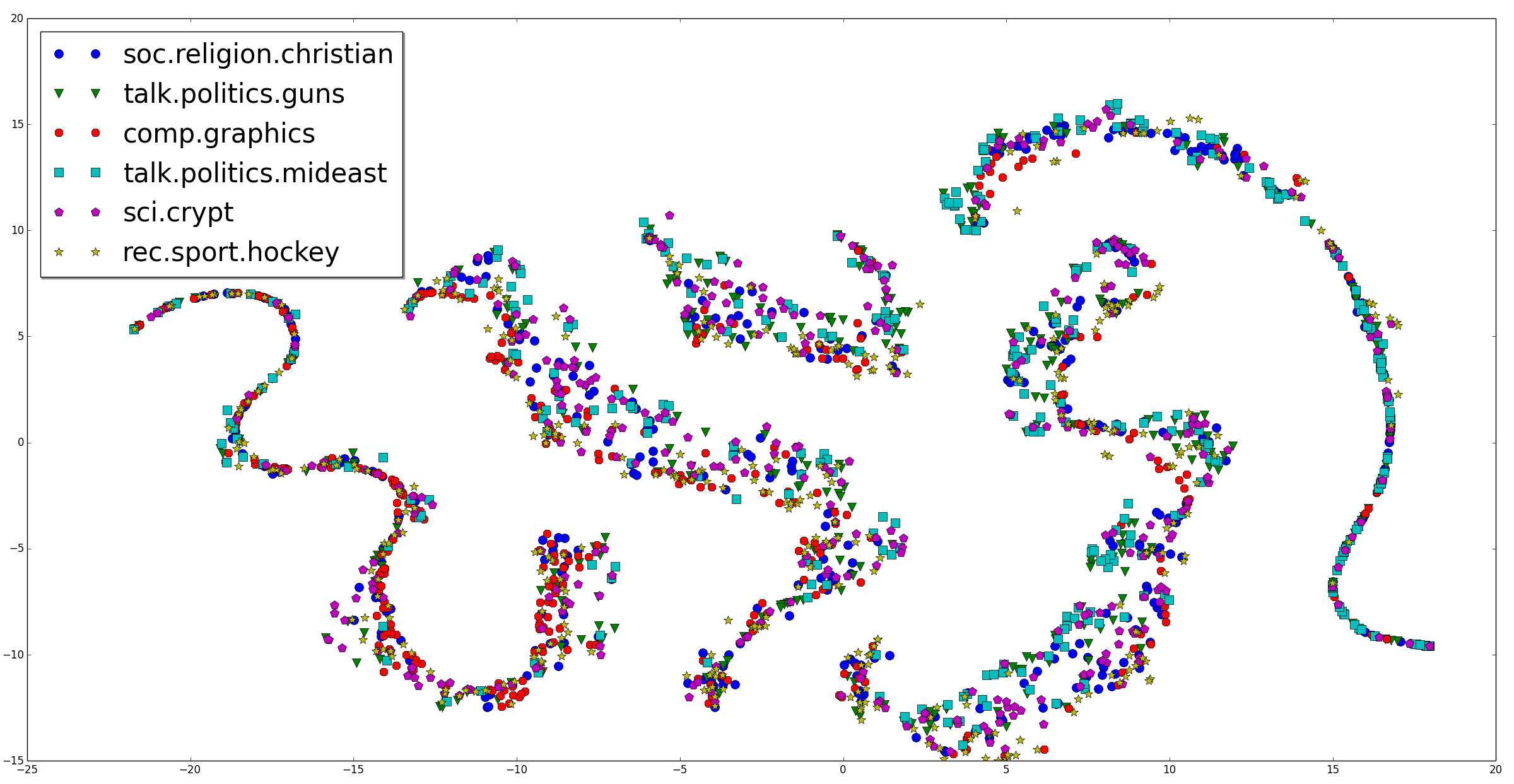}
	\label{fig:TSNE_AE}
	}
  \subfloat[t][KSAE]{%
    \includegraphics[width=1.75in, height=1.5in]{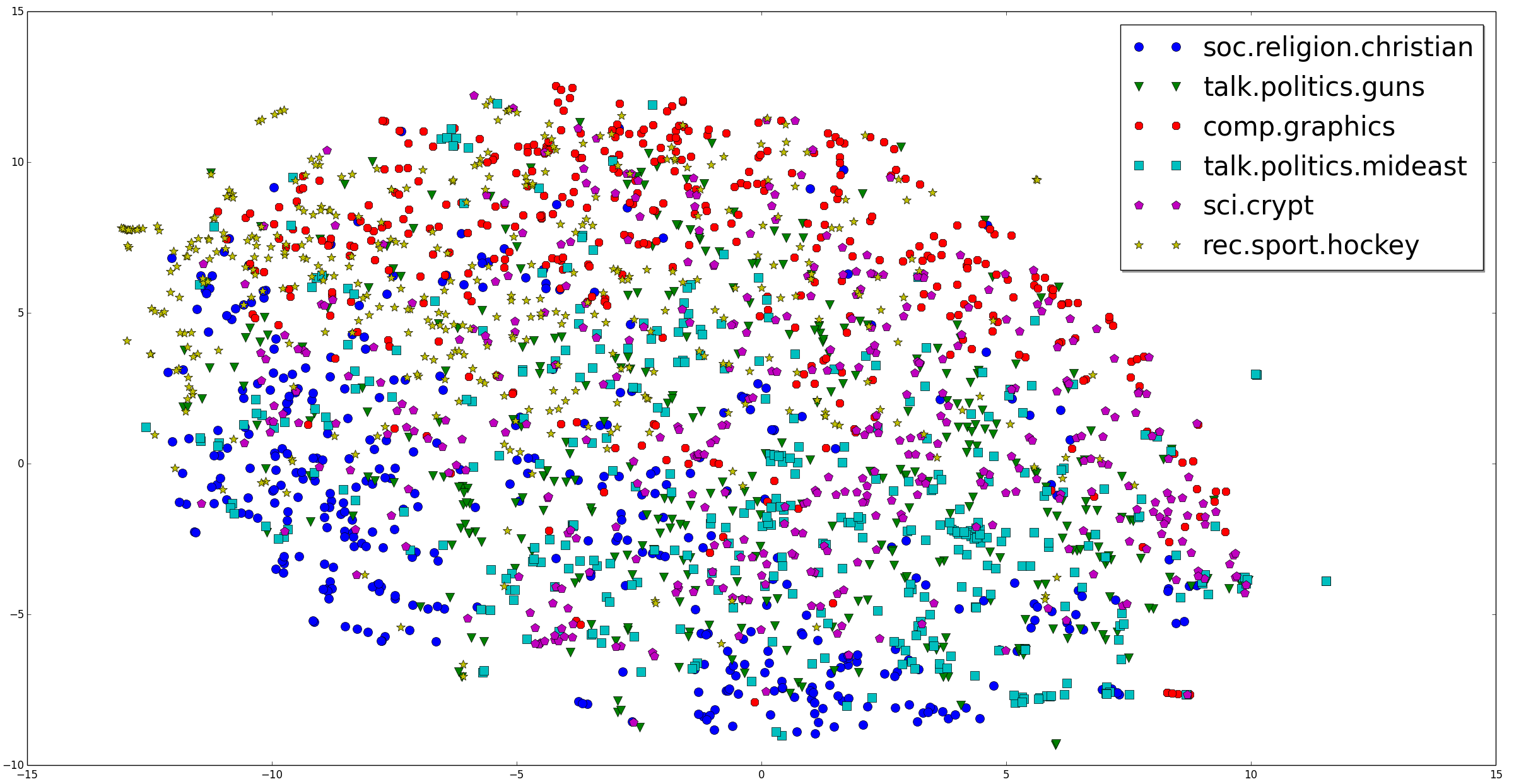}
    \label{fig:TSNE_KSAE}
    }
  \subfloat[t][LDA]{%
    \includegraphics[width=1.75in, height=1.5in]{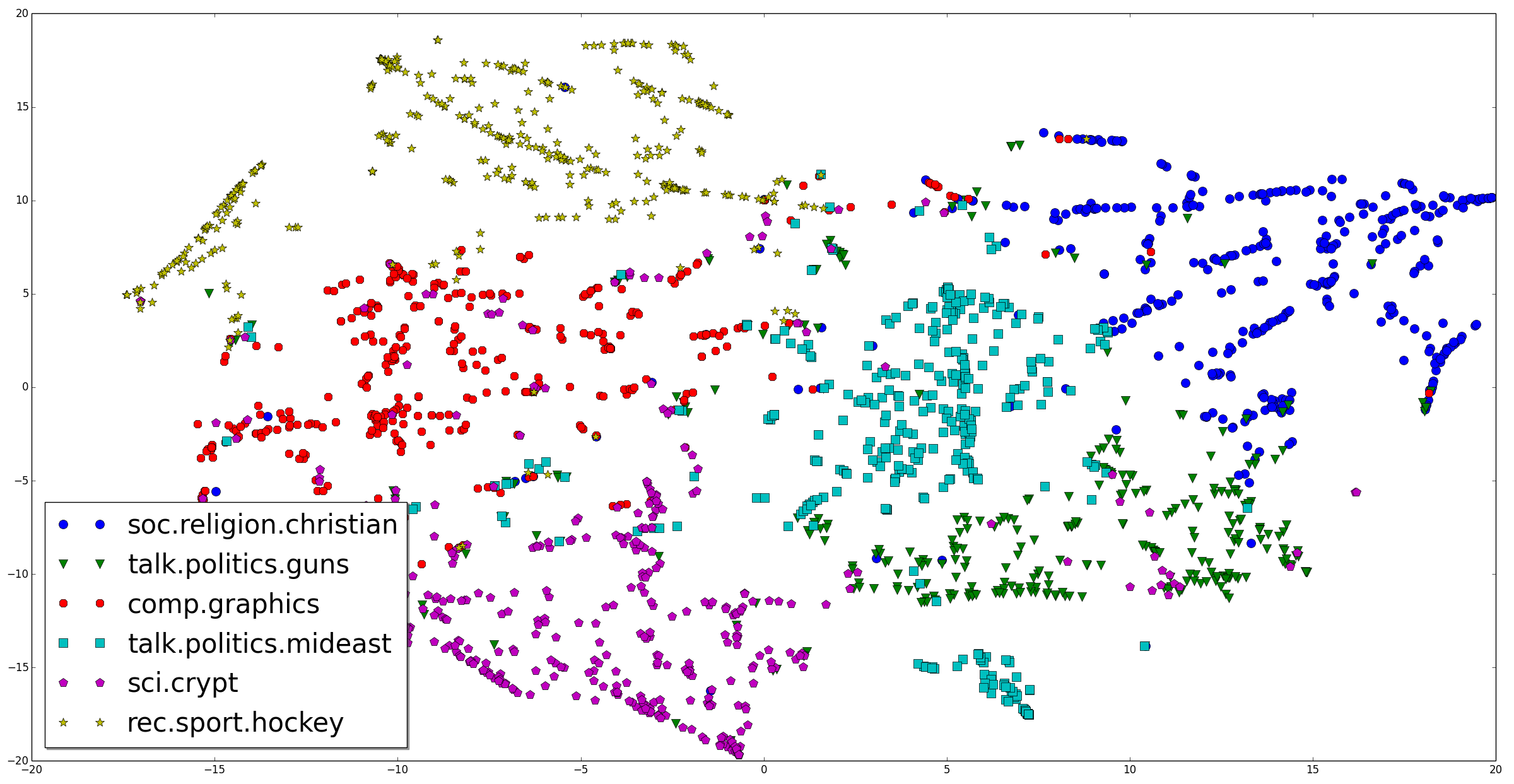}
    \label{fig:TSNE_LDA}
    }
  \subfloat[t][\alg]{%
    \includegraphics[width=1.75in, height=1.5in]{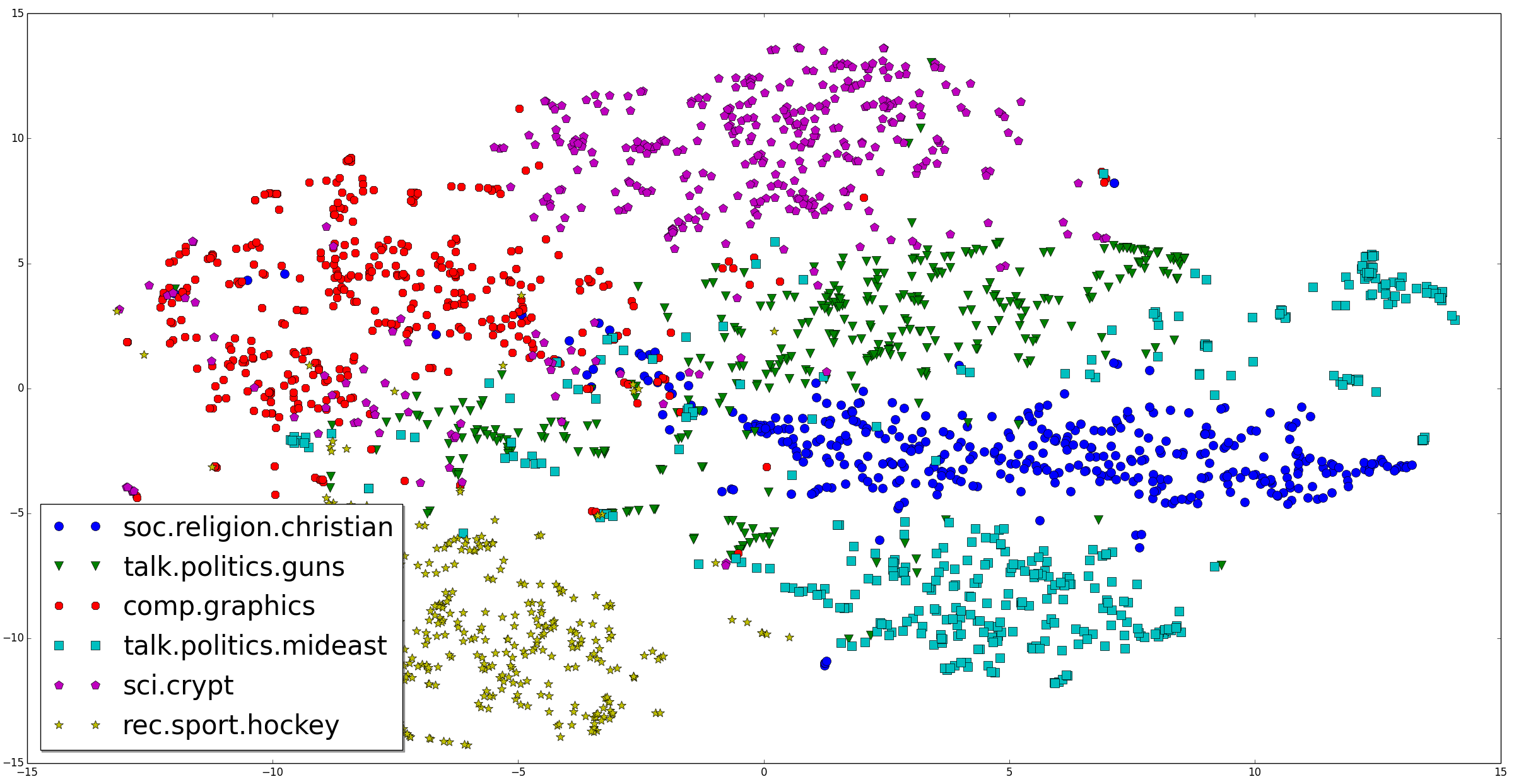}
    \label{fig:TSNE_KATE}
   }}
  \vspace{-0.1in}
  \caption{T-SNE on the 20-dimensional document vectors from 20
      Newsgroups dataset.}
  \label{fig:TSNE}
  \vspace{-0.2in}
\end{figure*}

\subsubsection{Word Embeddings Learned by Different Models}

In AE, KSAE and \alg, each input neuron (i.e., a word in the vocabulary set) is
connected to each hidden neuron (i.e., a virtual topic) with different
strengths. Thus, each row $i$ of the input to hidden layer weight matrix
$\bW \in \setR^{d\times m}$ is taken as an $m$-dimensional word
embedding for word $i$. In order to evaluate whether \alg can capture
semantically meaningful word representations, we check if similar or
related words are close to each other in the vector space. 
Table~\ref{table:similar_words} shows the five nearest neighbors for some
query words in the word representation space learned by AE, KSAE, \alg and Word2Vec. 
\alg performs much better than AE and KSAE. 
For example \alg lists words like {\em arm, crime, gun, firearm,
    handgun} among the nearest neighbors of query word {\em weapon}
    while neither AE nor KSAE is able to find relevant words.
One can observe that \alg can learn competitive word representations
compared to Word2Vec in terms of this word similarity task.

\subsubsection{Visualization of Document Representations}

A good document representation method is expected to group related
documents, and to separate the different groups.
Figure~\ref{fig:PCA} shows the PCA projections of the document
representations taken from the six main groups
in the 20 Newsgroups data.
As we can observe, neither AE nor the KSAE methods can learn good
document representations.
On the other hand, \alg successfully extracts meaningful representations 
from the documents; it automatically clusters related documents in the
same group, and it can easily distinguish the six different groups. 
In fact, \alg is very competitive with LDA (arguably even better on this
dataset, since
LDA confuses some categories), even though the latter
explicitly learns documents representations as mixture of topics, which
in turn are mixture of words.
Figure~\ref{fig:TSNE} shows the T-SNE based
visualization~\citep{maaten2008visualizing} of the above document
representations and we can draw a similar conclusion.

\subsection{Quantitative Experiments}
We now turn to quantitative experiments to measure the effectiveness of
\alg compared to other models on tasks such as classification,
multi-label classification (MLC), regression, and document retrieval
(DR).
For classification, MLC and regression tasks, we train a simple neural
network that uses the encoded test inputs
as feature vectors, and directly maps them to the output
classes or values. 
A simple softmax classifier with cross-entropy loss was applied for the
classification task, and multi-label
logistic regression 
classifier with cross-entropy loss was applied for the MLC task.
For the regression task we used a two-layer neural regression model
(where the output layer is a sigmoid neuron) with squared error loss. 
The same architecture is used for all methods to ensure fairness.
Note that when comparing various methods, the same number of features were learned for all of them except for
Word2Vec$_{pre}$ which uses 300-dimensional pre-trained word embeddings
and thus its number of features was fixed as 300 in all experiments.

\subsubsection{Mean Squared Cosine Deviation among Topics}\label{sec:mscd}

We first quantify how distinct are the topics learned via different
methods. 
Let $\bv_i$ denote the 
vector representation of topic $i$, and let there be $m$ topics.
The cosine of the angle between $\bv_i$
and $\bv_j$, given as $\cos(\bv_i, \bv_j) =
\tfrac{\bv_i^T\bv_j}{\|\bv_i\| \cdot \|\bv_j\|}$, 
is a measure of how similar/correlated the two topic vectors are; it
takes values in the range $[-1,1]$.
The topics are most dissimilar when the vectors are orthogonal to each
other, i.e., with the angle between them is $\pi/2$, with the cosine of
the angle being zero. 
Define the pair-wise mean squared cosine deviation among $m$ topics as
follows
$$MSCD = \sqrt{\frac{2}{m(m-1)}\sum_{i, j > i}
    {\cos^2(\bv_i, \bv_j)}}$$
Thus, MSCD $ \in [0,1]$, and smaller values of MSCD (closer to zero) 
imply more distinctive, i.e., orthogonal topics.

\begin{table}[!ht]
    \small
 \begin{minipage}{0.5\textwidth}
  \centering
    \begin{tabular}{|p{1.92cm}|c|c|c|}
    \hline
    \multirow{2}{.2cm}{\textbf{Model}} & \multicolumn{3}{c|}{\textbf{20 News}}\\
    \cline{2-4}
    & \textbf{20} & \textbf{128} & \textbf{512}\\
    \hline
   AE & 0.976 & 0.722   &  0.319 \\  
   KSAE & 0.268 &  0.198  & 0.056  \\  
   LDA & 0.249  &  0.059  &  0.028 \\ 
   \alg$_{no\_amp.}$ &  0.154  & 0.069  &  0.037\\ 
   \alg &  \textbf{0.097}  & \textbf{0.024}  & \textbf{0.014} \\   
  \hline
  \end{tabular}
  \caption{Mean squared cosine deviation among topics; smaller means
      more distinctive topics.}
  \label{table:mean_squared_deviation}  
  \end{minipage}
  \vspace{-0.2in}
\end{table}

We evaluate MSCD for topics generated by AE, KSAE, LDA, and \alg.
We also evaluate \alg without amplification.
In LDA, a topic is represented as its probabilistic
distribution over the vocabulary set, whereas for autoencoders,
it is defined as the weights on the connections between the corresponding hidden
neuron and all the input neurons.
We conduct experiments on the 20 Newsgroups dataset and
vary the number of topics from 20 to 128 and 512.
Table~\ref{table:mean_squared_deviation} shows these results.
We find that \alg has the lowest MSCD values, which means that it 
can learn more distinctive (i.e., orthogonal) topics than other
methods. Our results are much better than LDA, since the latter does not
prevent topics from being similar. On the other hand, the competition in
\alg drives topics (i.e., the hidden neurons) to become distinct from each other.
Interestingly, \alg with amplification (i.e., here we have $\alpha=6.26$) consistently achieves lower MSCD values 
than \alg without amplification, which verifies 
the effectiveness of the $\alpha$ amplification connections in terms of
learning distinctive topics.

\begin{table}[!ht]
\centering
\small
  \centering
  \renewcommand{\arraystretch}{1.0}
  \begin{tabular}{|p{1.8cm}|c|c|}
    \hline
    \multirow{2}{.2cm}{\textbf{Model}} & \multicolumn{2}{c|}{\textbf{20 News}}\\
    & \textbf{128} & \textbf{512}\\
    \hline
   LDA & 0.657   & 0.685 \\  
   DBN & 0.677  &   0.705  \\ 
   DocNADE &  0.714  &  0.735 \\ 
   NVDM &  0.052  &  0.053  \\ 
   Word2Vec$_{pre}$ & 0.687  & 0.687  \\ 
   Word2Vec  &  0.564 & 0.586   \\  
   Doc2Vec &  0.347   &  0.399   \\ 
   
   AE & 0.084  & 0.516  \\ 
   DAE &0.125   & 0.291  \\  
   CAE & 0.083 & 0.512  \\ 
  VAE & 0.724 & 0.747 \\ 
  KSAE & 0.486 & 0.675 \\ 
  \alg &\textbf{0.744}  &  \textbf{0.761} \\  
  \hline\end{tabular}
  \caption{Classification accuracy on 20 Newsgroups dataset.}
  \label{table:mcc_results}
\vspace{-0.2in}
\end{table}

\subsubsection{Document Classification Task}

In this set of experiments, we evaluate the quality of learned document
representations from various models for the purpose of document
classification. Table~\ref{table:mcc_results} shows the classification
accuracy results on the 20 Newsgroups dataset (using 128 topics). 
Traditional autoencoders
(including AE, DAE, CAE) do not perform well on this task. We observed
that the validation set error was oscillating when
training these classifiers (also observed in the regression task below), 
which indicates that the extracted features are
not representative and consistent. KSAE consistently achieves
higher accuracies than other autoencoders and does not
exhibit the oscillating phenomenon, which means that adding
sparsity does help learn better representations. 
VAE even performs better than KSAE on this dataset, which shows the
advantages of VAE over other traditional autoencoders.
However, as we will see later, VAE fails to consistently perform well across different datasets and tasks. 
Word2Vec$_{pre}$ performs on par with DBN and LDA even though it just averages all
the word embeddings in a document, which suggests the effectiveness of
pre-training word embeddings on a large external corpus to learn
general knowledge. 
Not surprisingly, DocNADE works very well on this
task as also reported in previous work~\citep{larochelle2012neural,
    srivastava2013modeling, cao2015novel}.
Our \alg model significantly outperforms all other models. For
example, \alg obtains
74.4\% accuracy which is significantly higher than the 72.4\% accuracy
achieved by VAE.

Table~\ref{table:reuters_mlc_results} shows multi-label
classification results on Reuters and Wiki10+ datasets. Here we show
both the Macro-F1 and Micro-F1 scores (reflecting a balance of precision
and recall) for different number of features.
Micro-F1 score biases the metric towards the most populated labels,
while Macro-F1 biases the metric towards the least populated
labels. 
Both Reuters and Wiki10+ are highly imbalanced. For example in
Wiki10+, the documents belonging to `wikipedia' or `wiki'
account for 90\% of the corpus while only around 6\% of the documents
are relevant to `religion'. Similarly, in Reuters, the
documents belonging to `CCAT' account for 47\% of the corpus while
there are only 5 documents relevant to `GMIL'.
DocNADE works the very well on this task, but the sparse and competitive
autoencoders also perform well. \alg outperforms KSAE on Reuters and
remains competitive on Wiki10+.
We don't report the results of DBN on Reuters since the training did not
end even after a long time.

\begin{table*}[!ht]
  \vspace{-0.1in}
  \centering
  \small
  \begin{tabular}{|p{1.8cm}||c|c|c|c||c|c|c|c|}
    \hline
    \multirow{3}{.2cm}{\textbf{Model}} & 
    \multicolumn{4}{c||}{\bf Reuters} & \multicolumn{4}{c|}{\bf Wiki10+}\\
    \cline{2-9}
    & \multicolumn{2}{c|}{\textbf{128}} & \multicolumn{2}{c||}{\textbf{512}} &
    \multicolumn{2}{c|}{\textbf{128}} & \multicolumn{2}{c|}{\textbf{512}} \\
    & \textbf{Macro-F1} & \textbf{Micro-F1} & \textbf{Macro-F1} &
    \textbf{Micro-F1} &
    \textbf{Macro-F1} & \textbf{Micro-F1} & \textbf{Macro-F1} & \textbf{Micro-F1}\\
    \hline
   LDA & 0.408 & 0.703 & 0.576  & 0.766 &  
   0.442 & 0.584   &  0.305 &  0.441 \\  
   DBN &- &- &- &- & 
   0.330  &  0.513  &  0.339  & 0.536  \\ 
   DocNADE &   {\bf 0.564} &  {\bf 0.768}  &   {\bf 0.667} &  {\bf 0.831}  & 
   {\bf 0.451}  & 0.585 & 0.423 & 0.561 \\ 
   NVDM & 0.215  & 0.441 &  0.195 & 0.452 &  
   0.187  & 0.461 &0.036  & 0.375 \\ 
   Word2Vec$_{pre}$ & 0.549   &  0.712  &  0.549  &  0.712 & 
   0.312  & 0.454 & 0.312  & 0.454 \\ 
   Word2Vec  & 0.458 & 0.648 & 0.595 &  0.761 & 
   0.205 & 0.318  & 0.234    & 0.325 \\ 
   Doc2Vec & 0.004 & 0.082 & 0.000 &  0.000 & 
   0.289   & 0.486 & 0.344  & 0.524   \\   
   AE & 0.025  & 0.047  & 0.459   & 0.651 & 
   0.016  & 0.040   & 0.382  &  0.569\\ 
   DAE & 0.275 & 0.576 &   0.489 & 0.685 & 
   0.359 &  0.560  & 0.375  & 0.534\\  
   CAE & 0.024 & 0.045  & 0.549 &  0.726 & 
   0.091   &  0.168  & 0.404  & 0.547  \\ 
   VAE & 0.325 & 0.458 &   0.490 & 0.594 & 
   0.342 & 0.497  & 0.373 & 0.511 \\ 
   KSAE &  0.457   &    0.660 & 0.605 & 0.766  & 
    0.449  & {\bf 0.594} &  {\bf 0.471}   &  {\bf 0.614}  \\ 
   \alg & 0.539  &  0.716 &  0.615  & 0.767 & 
  0.445 & 0.580  &0.446 &  0.580\\ 
  \hline\end{tabular}
   \captionsetup{justification=centering}
  \caption{Comparison of MLC F1 score on Reuters RCV1-v2 and Wiki10+ datasets.}
  \label{table:reuters_mlc_results}
  \label{table:wiki_mlc_results}
  \vspace{-0.1in}
\end{table*}

\begin{table}[!h]
  \centering
  \small
    \begin{tabular}{|p{1.8cm}|c|c|}
    \hline
    \multirow{2}{.2cm}{\textbf{Model}} & \multicolumn{2}{c|}{\textbf{MRD}}\\
    & \textbf{128} & \textbf{512}\\
    \hline
   LDA & 0.287  & 0.226 \\  
   DBN &  0.277  & 0.369 \\ 
   DocNADE &0.404  &  0.424  \\ 
   NVDM &  0.199  &  0.191\\ 
   Word2Vec$_{pre}$ & 0.409  & 0.409  \\ 
    Word2Vec &  0.143  &  0.136  \\   
    Doc2Vec & 0.052  & 0.032 \\ 
   AE &  -0.001  & 0.203 \\ 
   DAE & 0.067 & 0.100 \\ 
   CAE & 0.018  & 0.118 \\ 
  VAE & 0.111 & 0.355   \\ 
  KSAE &  0.152  & 0.365  \\ 
   \alg & \textbf{ 0.463 } & \textbf{0.516 }\\  
  \hline
  \end{tabular}
  \caption{Comparison of regression $r^2$ score on MRD dataset.}
  \label{table:regression_results}
  \vspace{-0.1in}
\end{table}

\subsubsection{Regression Task}

In this set of experiments, we evaluate the quality of learned document
representations from various models for predicting the movie ratings in
the MRD dataset, as shown in Table~\ref{table:regression_results} (using
128 features). 
The coefficient of determination, denoted $r^2$, from the regression
model was
used to evaluate the methods. The best possible $r^2$ statistic value is
1.0; negative values are also possible, indicating a poor fit of the
model to the data. 
In general, other autoencoder models perform poorly on this task, for example, 
AE even gets a negative $r^2$ score.
Interestingly, Word2Vec$_{pre}$ performs on par with DocNADE, indicating
that word embeddings learned from a large external corpus
can capture some semantics of emotive words (e.g., good, bad,
wonderful).
We observe that \alg significantly
outperforms all other models, including Word2Vec$_{pre}$, which means it 
can learn meaningful 
representations which are helpful for sentiment analysis.

\begin{figure}[!ht]
   \vspace{-0.1in}
     \includegraphics[width=3.5in,height=2in]{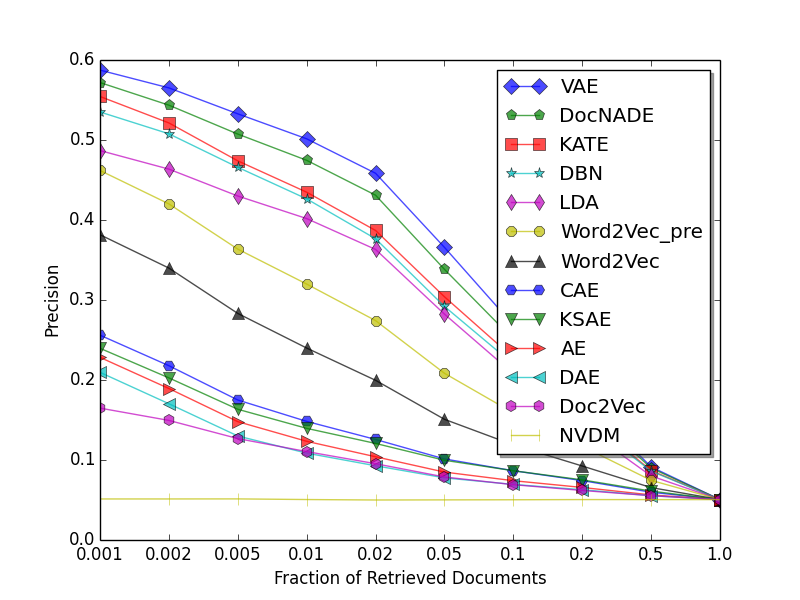}
     \vspace{-0.1in}
   \caption{Document retrieval on 20 Newsgroups dataset (128 features).}
   \label{fig:20News_Retrieval}
   \vspace{-0.2in}
\end{figure}

\subsubsection{Document Retrieval Task}


We also evaluate the various models for document
retrieval. Each document in the test set is used as an individual query and
we fetch the relevant documents from the training set based on the cosine
similarity between the document representations. The average fraction of
retrieved documents which share the same label as the query document,
i.e., precision, was used as the evaluation metric.
As shown in Figure~\ref{fig:20News_Retrieval}, VAE performs the best
on this task followed by DocNADE and \alg. Among the other models, DBN and LDA also
have decent performance, but the other autoencoders are not that
effective.

\subsubsection{Timing}

\begin{table*}[!ht]
    \small
  \centering
    \begin{tabular}{|c|c|c|c|c|c|c|c|c|c|c|c|c|}
    \hline
   Model & LDA & DBN & DocNADE & NVDM & Word2Vec & Doc2Vec & AE & DAE & CAE & VAE & KSAE & \alg \\ \hline
      Time (s) & 399 & 15,281&4,787 &645 &977 & 992 & 566&361 &729 & 660 & 489 &1,214 \\ \hline
  \end{tabular}
  \caption{Training time of various models (in seconds).}
  \label{table:training_time}  
  \vspace{-0.1in}
\end{table*}


Finally, we compare the training time of various models. Results are
shown in Table~\ref{table:training_time} for the 20 Newsgroups dataset,
with 20 topics.
Our model is much faster than deep generative models like DBN and
DocNADE. It is typically slower than other autoencoders since it 
usually takes more epochs to converge. Nevertheless, as demonstrated
above, it significantly
outperforms other models in various text analytics tasks.


\begin{figure*}[!h]
   \vspace{-0.1in}
  \centerline{
  \subfloat[t][Effect of number of topics]{%
      \label{fig:effects_topics}
    \includegraphics[width=2in]{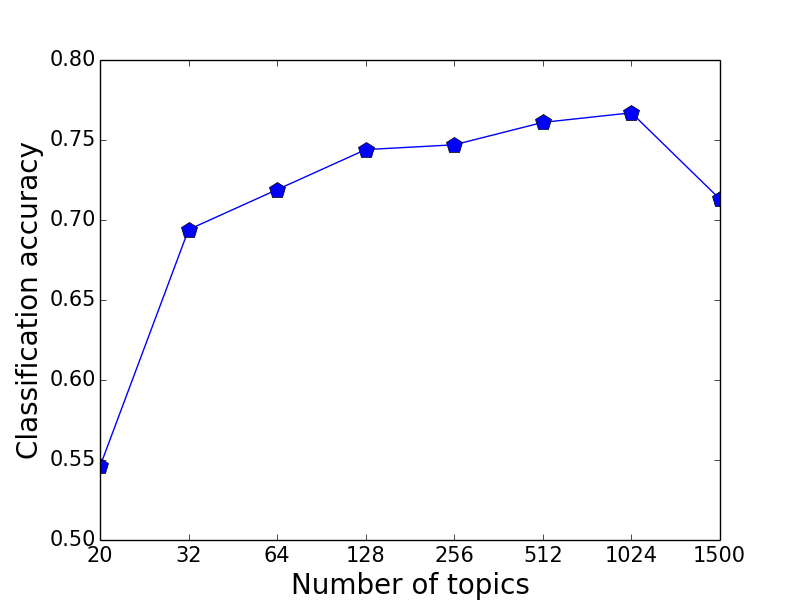}
 }
  \subfloat[t][Effect of $k$]{%
      \label{fig:effects_k}
    \includegraphics[width=2in]{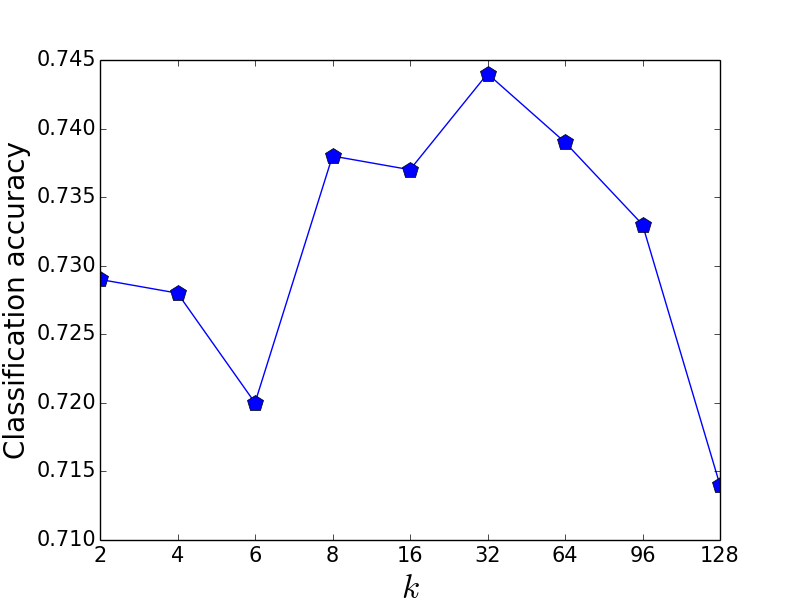}
  }
   \subfloat[t][Effect of $\alpha$]{%
      \label{fig:effects_alpha}
    \includegraphics[width=2in]{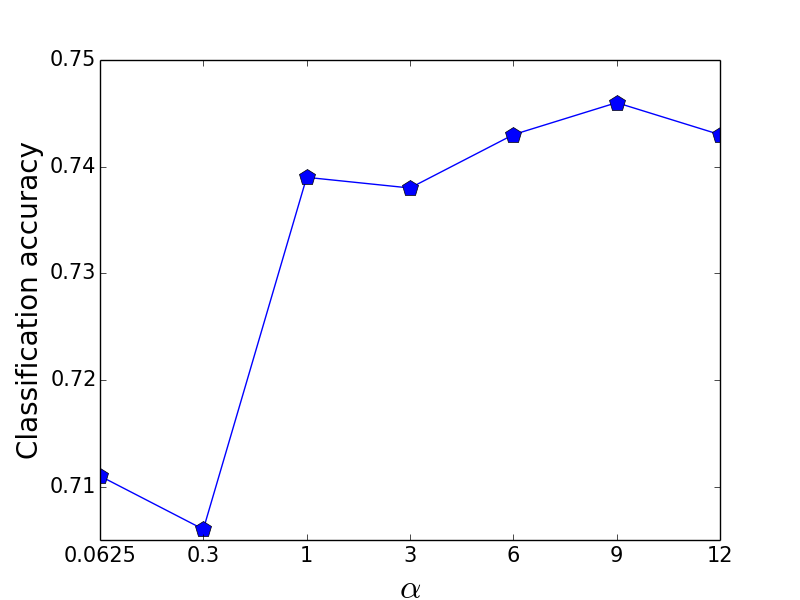}
    }
    }
   \vspace{-0.1in}
   \caption{Effects of hyper-parameters.}
   \label{fig:effects_hyperparam}
   \vspace{-0.1in}
\end{figure*}

\subsection{\alg: Effects of Parameter Tuning}\label{sec:effects_of_params}
Having demonstrated the effectiveness of \alg compared to other methods,
we study the effects of various hyperparameter choices in \alg, such as
the number of topics (i.e., hidden neurons), the number of winners $k$
and the energy amplification parameter $\alpha$.
The default values for the number of topics is 128, with $k=32$ and
$\alpha=6.26$.
Note when exploring the effect of the number of topics, 
we also vary $k$ to find its best match to the given number of topics.
Figure~\ref{fig:effects_hyperparam} shows the classification accuracy
on the 20 Newsgroups dataset, as we vary these parameters.
We observe that as we increase the number of topics or hidden neurons
(in Figure~\ref{fig:effects_topics}),
the accuracy continues to rise, but eventually drops off. We use 128 as
the default value since it offers the best trade-off in complexity and
performance; only relatively minor gains are achieved in increasing the
number of topics beyond 128. Considering the number of winning neurons
(see Figure~\ref{fig:effects_k}),
the main trend is that the performance degrades when we make
$k$ larger, which is expected since larger $k$ implies lesser competition.
In practice, when tuning $k$, we find that starting by a value close to around 
a quarter of the number of topics is a good strategy.
Finally, as we mentioned, the $\alpha$ amplification connection
is crucial as verified in Figure~\ref{fig:effects_alpha}.
When $\alpha=2/k=0.0625$, which means there is no amplification for the energy,
the classification accuracy is 71.1\%. However, we are able to
significantly boost the model performance up to 74.6\% accuracy by increasing the value of $\alpha$. 
We use a default value of $\alpha=6.26$, which once again reflects a
good trade-off across different datasets.
It is also important to note that across all the experiments, we found that
using the {\em tanh} activation function (instead of {\em sigmoid}
function) in the k-competitive layer of \alg gave the best performance.
For example, on the 20 Newsgroups data, using 128 topics,
\alg with {\em tanh} yields 74.4\% accuracy, while with {\em sigmoid} it
was only 56.8\%.

\section{Conclusions}
We described a novel k-competitive autoencoder, \alg, that explicitly
enforces competition among the neurons in the hidden layer by selecting
the $k$ highest activation neurons as winners, and reallocates
the amplified energy (aggregate activation potential) from the losers.
Interestingly, even though we use a shallow model, i.e., with one hidden
layer, it outperforms a variety of methods on many different text
analytics tasks. More specifically, we perform a comprehensive
evaluation of \alg against techniques spanning graphical models (e.g.,
LDA), belief
networks (e.g., DBN), word embedding models (e.g., Word2Vec), and
several other autoencoders including the k-sparse autoencoder (KSAE). We
find that across tasks such as document classification, multi-label
classification, regression and document retrieval, \alg clearly
outperforms competing methods or obtains close to the best results. It is
very encouraging to note that \alg is also able to learn semantically
meaningful representations of words, documents and topics, which we
evaluated via both quantitative and qualitative studies. As part of
future work, we plan to evaluate \alg on more domain specific datasets,
such as bibliographic networks, for example 
for topic induction and scientific publication retrieval. We also plan
to improve the scalability and effectiveness of our approach on much
larger text collections by developing parallel and distributed
implementations.



\begin{acks}
This work was supported in part by \grantsponsor{}{NSF}{} awards
\grantnum{}{IIS-1302231} and \grantnum{}{IIS-1738895}.
    
\end{acks}

%
\bibliographystyle{ACM-Reference-Format}
\bibliography{reports}  
%
%

\end{document}